# Continual-learning-based framework for structural damage recognition

Jiangpeng Shu[1], Jiawei Zhang[*1,2], Reachsak Ly[1], Fangzheng Lin[3], Yuanfeng Duan[1]

*[1]College of Civil Engineering and Architecture, Zhejiang University, 310058 Hangzhou, China*

*[2]Center for Balance Architecture, Zhejiang University, 310058 Hangzhou, China*

*[3]Institute of Construction Informatics, Technische Universität Dresden, 01069 Dresden, Germany*

**Abstract:** *Multi-damage is common in reinforced concrete structures and leads to the requirement of large number of neural networks, parameters and data storage, if convolutional neural network (CNN) is used for damage recognition. In addition, conventional CNN experiences catastrophic forgetting and training inefficiency as the number of tasks increases during continual learning, leading to large accuracy decrease of previous learned tasks. To address these problems, this study proposes a continual-learning-based damage recognition model (CLDRM) which integrates the "learning without forgetting" continual learning method into the ResNet-34 architecture for the recognition of damages in RC structures as well as relevant structural components. Three experiments for four recognition tasks were designed to validate the feasibility and effectiveness of the CLDRM framework. In this way, it reduces both the prediction time and data storage by about 75% in four tasks of continuous learning. Three experiments for four recognition tasks were designed to validate the feasibility and effectiveness of the CLDRM framework. By gradual feature fusion, CLDRM outperformed other methods by managed to achieve high accuracy in the damage recognition and classification. As the number of recognition tasks increased, CLDRM also experienced smaller decrease of the previous learned tasks. Results indicate that the CLDRM framework successfully performs damage recognition and classification with reasonable accuracy and effectiveness.*

## 1 INTRODUCTION

### 1.1 Background

Reinforced concrete (RC) civil structures gradually approach their design life expectancy, posing a risk to the safety structures and people (Liu et al. 2019). Therefore, it is necessary to effectively inspect damages of RC structures. Conventionally, different manual inspections methods were used. However, these methods have the disadvantages of unreliable inspection results and considerable time consumption (Cha et al. 2017). Owing to the rapid development of machine learning, which has shown excellent performance in image processing and computer vision in recent years, researchers and engineers are increasingly applying it for damage detection and structural assessment (DeVries et al., 2018, Spencer et al., 2019 & Chen et al., 2018). One of the well-known applications of deep learning in civil engineering is crack detection and recognition. For instance, Kim et al. (2019) used a convolutional neural network (CNN) to determine the existence and location of concrete cracks. Chen and Jahanshahi (2018) combined CNNs with naive Bayes data fusion to analyze individual video frames for crack detection.

The types of damage of RC structures are various. Thus, the demand for the recognition and classification of damage types is also on the rise. Moreover, it is becoming easier to retrieve more structural-damage-related information in the form of image data based on the technical development of deep learning and data acquisition. For this reason, numerous solutions have been explored. A region-based CNN (Faster-RCNN, Ren et al., 2017) was adopted to detect different damage types such as concrete cracks and steel corrosion with two levels (medium and high), including bolt corrosion and steel delamination (Cha et al., 2018). In addition, it was utilized to detect spalling and severe damage for exposed and buckled rebars (Ghosh et al., 2020). YOLO (You Only Look Once, Redmon et al., 2016), which is another region-based CNN, was used to detect multiple concrete bridge damage types such as pop-out or exposed rebars (Zhang et al., 2020). Based on ImageNet (Krizhevsky et al., 2017), Gao and Mosalam (2018) proposed structural ImageNet with four baseline recognition functions, i.e., component type determination, spalling condition check, damage level evaluation, and damage type determination. They employed transfer learning (TL) to prevent overfitting, along with two strategies, namely, feature extraction (Shin et al., 2016) and fine-tuning (Yosinski et al., 2014). (Gao et al. 2020) proposed a large-scale multi-attribute benchmark dataset of structural Images, namely, the Pacific Earthquake Engineering Research (PEER) Hub ImageNet ($\phi$-Net).Based on this dataset , (Gao et al. 2019) introduced a Generative Adversarial Network model, namely, Deep Convolutional Generative Adversarial Network (DCGAN) and propose a Leaf-Bootstrapping (LB) method to improve the performance of this DCGAN.

## 1.2 Multitask Damage Recognition

It is worth noting that with the continuous development of deep learning, the number of recognition tasks in damage detection is increasing. One of the common problems among the abovementioned applications is the limitation of the number of recognition tasks for training. For example, ten different neural networks are required for ten different recognition tasks. Hence, the number of parameters increases exponentially with the number of damage recognition tasks, thereby making the training process time/resource consuming. Additionally, these training methods do not utilize the feature correlation between similar tasks such as crack detection and spalling detection. To address this issue, one neural network can be trained simultaneously for multiple recognition tasks. The trained model can identify similar features and characteristics with improved accuracy (Li and Hoiem, 2018). Nevertheless, the similar characteristics of certain recognition tasks present a few drawbacks. For instance, the image characteristics of spalling condition check and damage level evaluation are similar to a certain extent. If a model only outputs one prediction, then there is a problem. Therefore, to prevent this confusion in training methods, fully connected layer is separated to output the results of different tasks. The two most common methods for this are simple sequence training (Sutskever et al., 2014) and joint training (Caruana, 1997). Simple sequence training begins by training a model for the first task. Then, more neurons are subsequently added in the tail part of a fully connected layer to learn the feature of the next task. The common problem of this method is the highly likely catastrophic forgetting, where a network tends to forget the feature learned in the old task after the training for the new task is finished. This significantly reduces the recognition accuracy for the first task. Joint training (Caruana, 1997) can be applied for simultaneously training multiple recognition tasks. However, the datasets of all previous tasks are required for the training of any new recognition task. These training methods have the following limitations:

1) Flaws appear when the number of training datasets between each task is extremely unbalanced; this causes poor recognition accuracy for the tasks with small samples. In the field of civil engineering, datasets are extremely expensive. There is a lack of particular types of images, which could result in poor performance for the corresponding recognition tasks. In addition, as the number of tasks increases, the recognition accuracy for each task becomes unpredictable or decreases.

2) The total amount of data required for simultaneous training increases with the number of recognition tasks. Therefore, considerable computing resources are required to perform training; otherwise, the problem of memory overflow occurs in GPUs during training.

3) Whenever a new task is added to a model, all data used in the training of the previous task are required again for retraining. These data are simultaneously used to retrain the model again. This not only causes the problem mentioned in 2) but also exponentially increases data storage and training time as the number of tasks increases. Furthermore, the assumption that the data from old tasks are always available at the time of training a new task is not always satisfied.

## 1.3 Continual Learning

Continual learning has been proposed as a revolutionary method (Parisi et al., 2019) for two purposes: (1) to perform multitask recognition using one CNN; (2) to maintain an optimal recognition accuracy for previously trained tasks with a low training cost. Continual learning is a generic method, which can be adopted in any CNN framework. Continual learning utilizes the features and knowledge learned from the recognition of previous tasks, facilitates training for next tasks, and maintains the features learned from previous tasks. This is achieved by limiting changes in parameters, i.e., adding a regularization term. For instance, Kirkpatrick et al.(2017) proposed elastic weight consolidation as a specific sequence training method that quantifies the importance of the weights of previous tasks, flexibly updates the parameters during the training of a new task, maintains the features learned in previous tasks, and prevents catastrophic forgetting. Zenke et al. (2017) introduced synaptic intelligence into artificial neural networks, where each synapse accumulates and exploits task-relevant information over time to rapidly store new knowledge without forgetting previous knowledge.

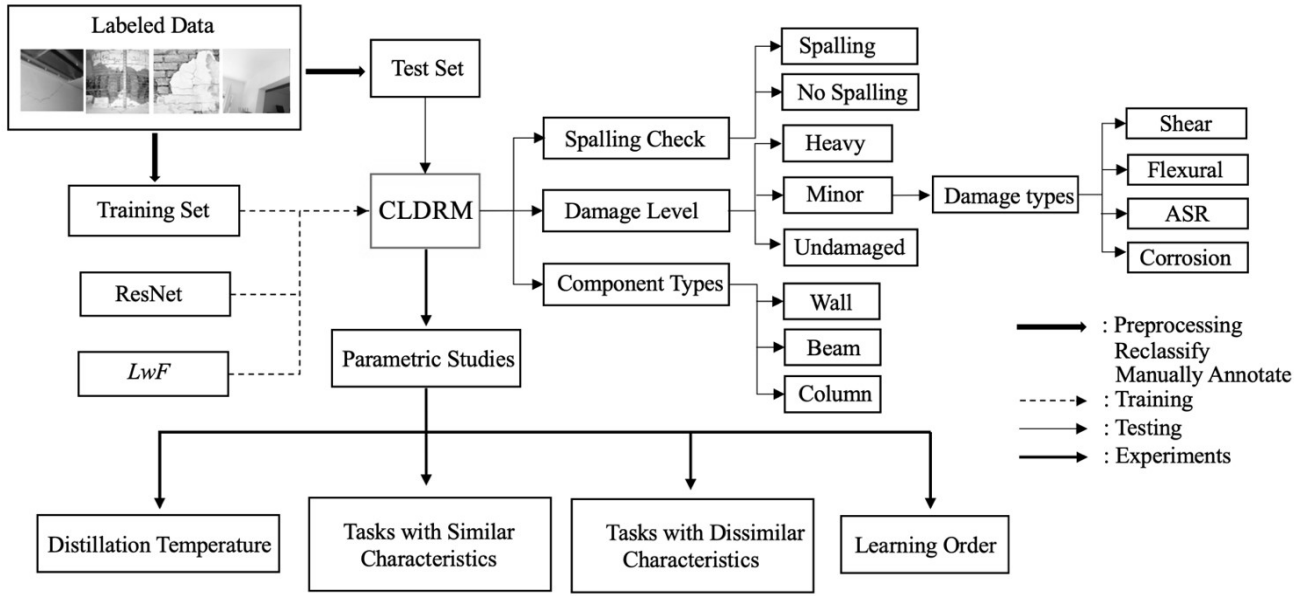

**Figure 1** Flowchart of continual-learning-based damage recognition model (CLDRM) framework.

In addition, a few rehearsal methods were developed to retrieve the knowledge of previous tasks. One classic example is deep generative replay, which is inspired by generative adversarial networks (Goodfellow et al., 2014). A set of separate generative models for previous tasks is trained to generate images for the subsequent training of a new task. However, considerable time and data storage are still required to train a network if the number of tasks is large.

To overcome the aforementioned limitations of previous methods, this study proposes a novel continual-learning-based damage recognition model (CLDRM), which can recognize multiple damage types of RC structures with the continual learning ability. The detailed objectives of this study are as follows: (1) To establish a database of multitask recognition, including damage level evaluation, spalling condition check, component type determination, and damage type determination, with approximately twenty thousand images. (2) To build a new deep CNN training framework, namely, the CLDRM, which utilizes the L*w*F method with the residual network (ResNet) (He et al., 2016) architecture for the detection and classification of crack damage in RC structures. (3) To study the effect of different parameters, such as distillation temperatures, correlation of features as well as learning orders, on the results of recognition based on proposed CLDRM.

## 2 OVERVIEW OF THE PROPOSED METHOD

Figure 1 shows the flowchart of the proposed framework, training steps, and testing steps. A comprehensive dataset that includes typical damage categories is prepared for training. A labeled training dataset is the input. A separate test dataset is prepared to verify the model and to investigate its performance on four different recognition tasks, which are (1) three-class classification for damage level evaluation, (2) binary classification for spalling condition check, (3) three-class classification for component type determination, and (4) four-class classification for damage type determination. Each task is defined in detail in Section 3.1. This study utilizes a collected large dataset of damaged RC image including PEER Hub ImageNet dataset (Gao and Mosalam, 2018). Overall, 16,840 colorful images with a resolution of $224 \times 224$ are prepared, preprocessed, reclassified, and manually annotated for the aforementioned recognition tasks. The hierarchy tree in Figure 2 illustrates the structure of the dataset. To address the lack of valuable structural images, each image is labeled with multiple tags as multiple attributes according to the recognition tasks. The final ratio of the images used in each recognition task is illustrated in Figure .

The CLDRM adopts the continual-learning-based "learning without forgetting (L*w*F)" method (Li and Hoiem, 2018), which is a form of knowledge distillation (Hinton et al. 2015). Instead of using a previous image dataset for retraining, the knowledge learned from previous tasks is retained by learning the soft targets provided by the previous model, which is stored after training. In other words, the previous model can be saved and stored for subsequent training without any image datasets of a previous task. This significantly reduces memory requirement and training complexity. A new deep CNN training framework, namely, the CLDRM, which utilizes the L*w*F method with the residual network (ResNet) (He et al., 2016) architecture for the detection and classification of crack damage in RC structures, is developed.

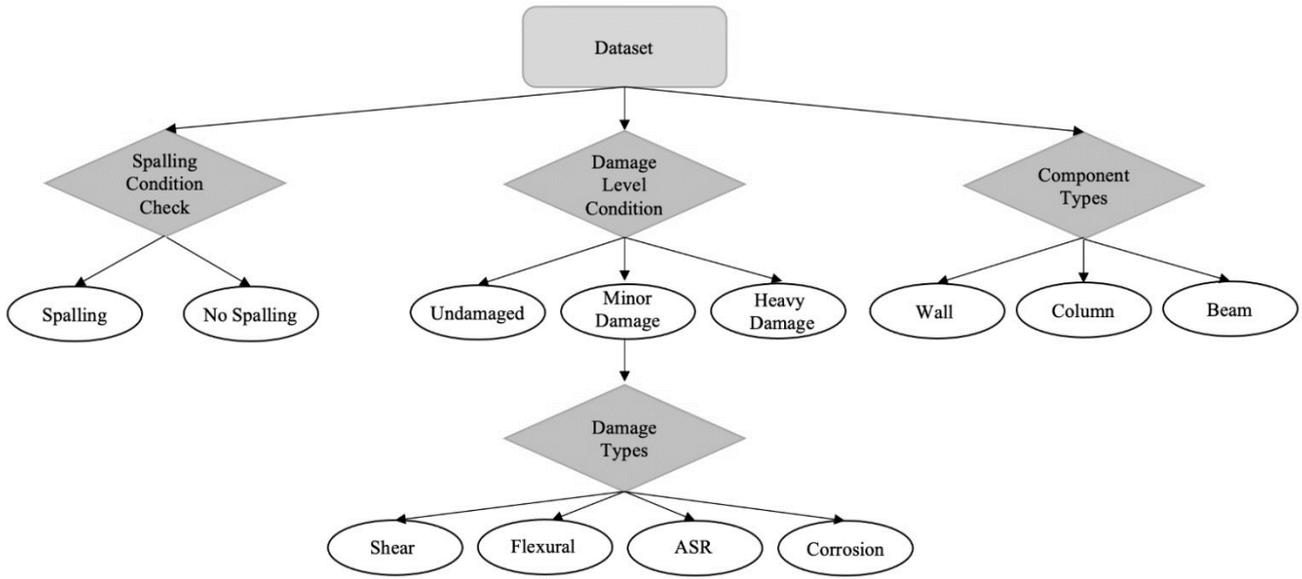

**Figure 2** Hierarchy tree of dataset used for CLDRM.

The effect of different parameters, such as distillation temperatures, correlation of features as well as learning orders, on the results of recognition based on proposed CLDRM, is investigated. This research's content is described as follows. Section 2 presents the synopsis of the proposed method. Section 3 provides the details of the four recognition tasks and their corresponding datasets. Section 4 explains the overall architecture of the networks and the proposed continual learning-based model training methodologies as well as the hyperparameters, dataset and settings used to train the model. Section 5 demonstrates how the framework evaluates test images from the proposed experiments and includes discussions regarding the method's performance and potential. Section 6 concludes this article and overviews some future possible studies to be conducted in this research area.

## 3 RECOGNITION TASKS AND DATASET

### 3.1 Recognition Tasks

#### 3.1.1 Damage level evaluation

Owing to limited datasets and engineering subjectivity on determining moderate versus heavy damage levels, previous researchers in this field have categorized the types of damage into three different levels, i.e., "undamaged," "minor damage," and "moderate damage to heavy damage." To achieve a clearer boundary between the levels of damage, this study categorizes crack damage into three main categories, i.e., "undamaged," "minor damage," and

"heavy damage," which are defined as follows: Minor damage does not pose a risk and does not need to be repaired urgently. Heavy damage can pose a risk and must be repaired urgently. The samples of the images used in training for damage level evaluation are shown in Figure 4.

#### 3.1.2 Spalling condition check

Spalling is the breaking of a concrete surface, and it frequently extends to the top layer of reinforcing steel. The spalling of concrete affects a broad variety of structures, including framed buildings, multistory car parks, and bridges. Spalling can be caused by the corrosion of embedded reinforcing steel, fire exposure, freeze and thaw cycling, and so on. Spalling is considered as a separate damage type category owing to its distinguishable characteristics (Figure ).

#### 3.1.3 Component type determination

Component type determination is a three-class classification task for "columns," "walls (or slabs)," and "beams" (Figure ). Owing to the geometrical similarity between beams and columns, rotation is not implemented for data augmentation to prevent the occurrence of label inconsistency between beams and columns.

#### 3.1.4 Damage type determination

Four typical crack damage types are selected based on the causes of cracks. All of them have evident characteristics, which are suitable for identification. The proportion of images for each crack type is shown in Figure 7.

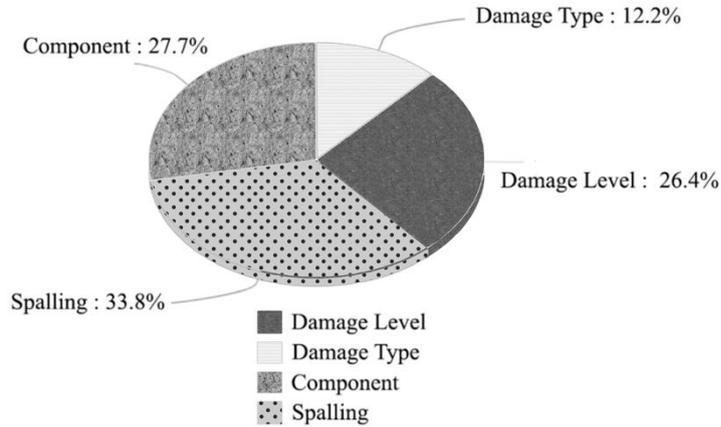

**Figure 3** Proportion of images used in different recognition task.

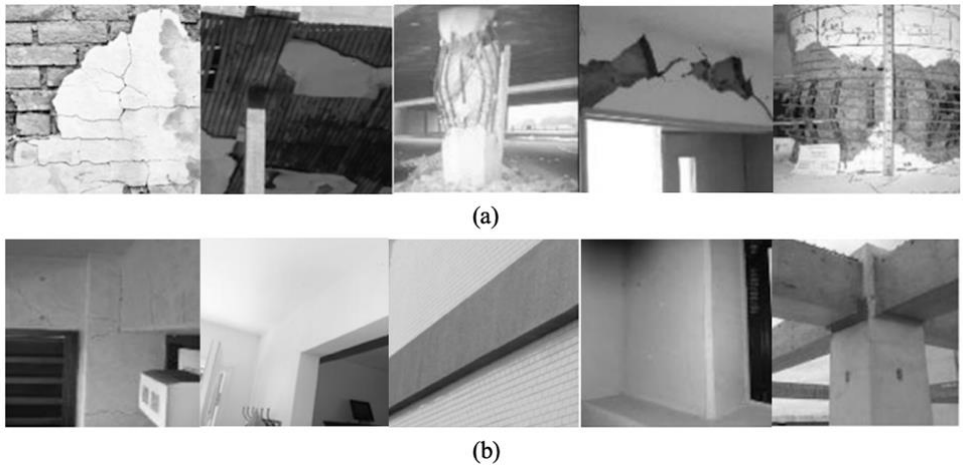

(a)

(b)

**Figure 3** Sample images used for damage level evaluation task: (a) No damage; (b) Minor damage; (c) Heavy damage.

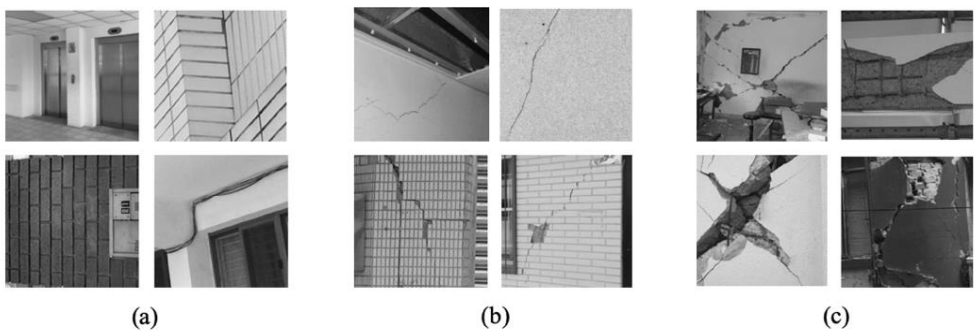

(a)                    (b)                    (c)

**Figure 5** Sample images used for spalling condition check: (a) Spalling; (b) No spalling.

1) Shear cracks: An extremely large shear stress causes cracks close to supports such as walls or columns. These cracks are typically inclined at 45°, and they have X or V shapes in a few cases (Figure ).

2) Flexural cracks: Flexural damage is caused by overloaded bending stress. It mostly occurs in the horizontal or vertical direction or at the end of a component with a horizontal or vertical edge.

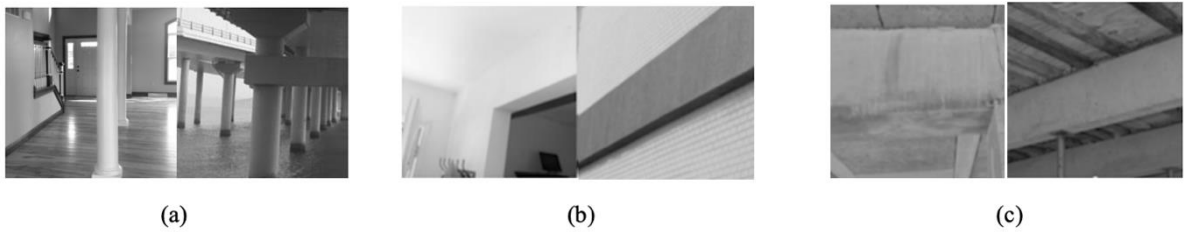

(a)                              (b)                              (c)

**Figure 6** Sample images used for component type determination: (a) Column; (b) Wall; (c) Beam.

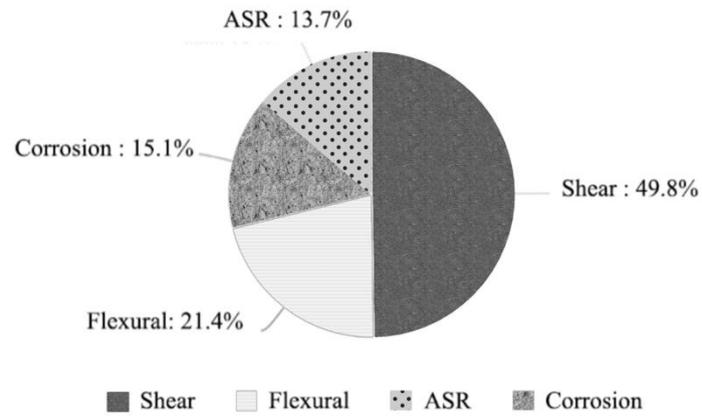

**Figure 4** Proportion of images used for each crack type.

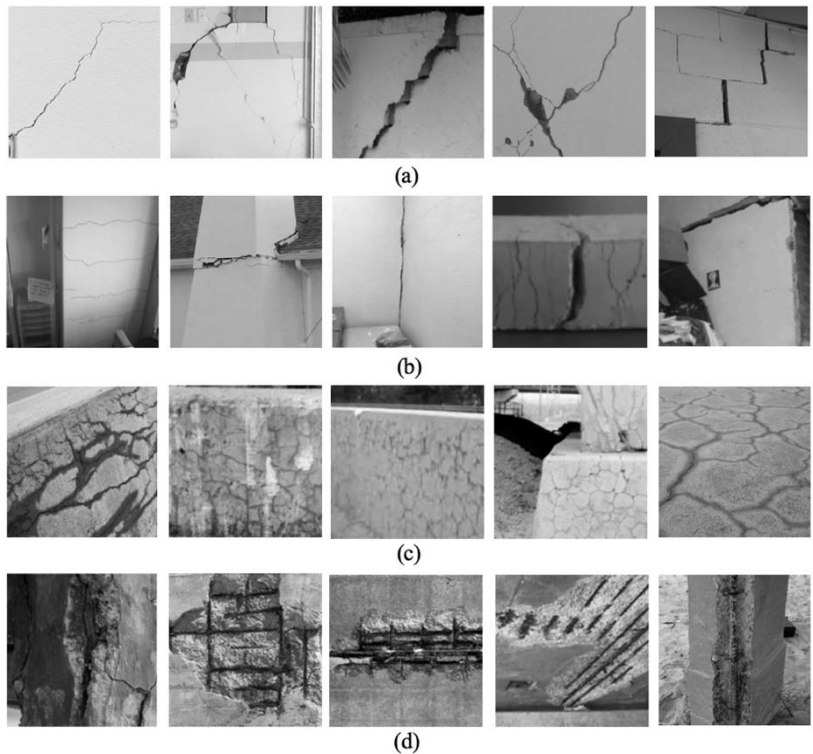

**Figure 8** Sample images used for damage type determination:
(a) Shear crack; (b) Flexural crack; (c) Cracks caused by Alkali–Silica
Reaction; (d) Cracks caused by corrosion

**Table 1** Variants of ResNet architecture (He et al., 2016).

| Block | Output size | 18-layer | 34-layer | 50-layer | 101-layer | 152-layer |
|---|---|---|---|---|---|---|
| Input | 224×224 | - | | | | |
| Conv Block 1 | 112×112 | 7×7, 64, stride 2 | | | | |
| Conv Block 2_x | 56×56 | 3×3 max pool, stride 2 | | | | |
| | | $\begin{bmatrix} 3 \times 3,64 \\ 3 \times 3,64 \end{bmatrix} \times 2$ | $\begin{bmatrix} 3 \times 3,64 \\ 3 \times 3,64 \end{bmatrix} \times 3$ | $\begin{bmatrix} 1 \times 1,64 \\ 3 \times 3,64 \\ 1 \times 1,256 \end{bmatrix} \times 3$ | $\begin{bmatrix} 1 \times 1,64 \\ 3 \times 3,64 \\ 1 \times 1,256 \end{bmatrix} \times 3$ | $\begin{bmatrix} 1 \times 1,64 \\ 3 \times 3,64 \\ 1 \times 1,256 \end{bmatrix} \times 3$ |
| Conv Block 3_x | 28×28 | $\begin{bmatrix} 3 \times 3,128 \\ 3 \times 3,128 \end{bmatrix} \times 2$ | $\begin{bmatrix} 3 \times 3,128 \\ 3 \times 3,128 \end{bmatrix} \times 4$ | $\begin{bmatrix} 1 \times 1,128 \\ 3 \times 3,128 \\ 1 \times 1,512 \end{bmatrix} \times 3$ | $\begin{bmatrix} 1 \times 1,128 \\ 3 \times 3,128 \\ 1 \times 1,512 \end{bmatrix} \times 4$ | $\begin{bmatrix} 1 \times 1,128 \\ 3 \times 3,128 \\ 1 \times 1,512 \end{bmatrix} \times 8$ |
| Conv Block 4_x | 14×14 | $\begin{bmatrix} 3 \times 3,256 \\ 3 \times 3,256 \end{bmatrix} \times 2$ | $\begin{bmatrix} 3 \times 3,256 \\ 3 \times 3,256 \end{bmatrix} \times 6$ | $\begin{bmatrix} 1 \times 1,256 \\ 3 \times 3,256 \\ 1 \times 1,1024 \end{bmatrix} \times 6$ | $\begin{bmatrix} 1 \times 1,256 \\ 3 \times 3,256 \\ 1 \times 1,1024 \end{bmatrix} \times 23$ | $\begin{bmatrix} 1 \times 1,256 \\ 3 \times 3,256 \\ 1 \times 1,1024 \end{bmatrix} \times 36$ |
| Conv Block 5_x | 7×7 | $\begin{bmatrix} 3 \times 3,512 \\ 3 \times 3,512 \end{bmatrix} \times 2$ | $\begin{bmatrix} 3 \times 3,512 \\ 3 \times 3,512 \end{bmatrix} \times 3$ | $\begin{bmatrix} 1 \times 1,512 \\ 3 \times 3,512 \\ 1 \times 1,2048 \end{bmatrix} \times 3$ | $\begin{bmatrix} 1 \times 1,512 \\ 3 \times 3,512 \\ 1 \times 1,2048 \end{bmatrix} \times 3$ | $\begin{bmatrix} 1 \times 1,512 \\ 3 \times 3,512 \\ 1 \times 1,2048 \end{bmatrix} \times 3$ |
| FC Block | 1×1 | average pool, 1000-d fc, softmax | | | | |

3) Cracks caused by Alkali–Silica reaction (ASR): These cracks (Grinys et al., 2014) occur over time in concrete between the highly alkaline cement paste and non-crystalline silicon dioxide found in common aggregates. This reaction can cause the expansion of the altered aggregate and leads to cracks and spalling.

4) Cracks caused by corrosion of reinforcing steel: Corrosion generally refers to stress corrosion cracking, which is induced by the combined influence of tensile stress and corrosive environments. Only extremely small concentrations of certain highly active chemicals can produce catastrophic cracking, which leads to devastating and unexpected failure

## 4 NETWORK ARCHITECTURE AND MODEL

### 4.1 Network Architecture

#### 4.1.1 Network Architecture of ResNet

ResNet (He et al., 2016) is well known for its remarkable object detection and image classification capabilities. It has been demonstrated that residual mapping and shortcut connections facilitate the training process and lead to better results compared to very deep plain networks. Network depth is crucial for improving the network performance of CNNs; however, deeper networks are more difficult to train. He et al. (2016) reported that an increase in network depth leads to degradation problems, which cause accuracy to become saturated. To address this problem, they used [batch normalization (BN), rectified linear unit (ReLU), and a convolution layer] × 2 as a basic mapping backbone and subsequently added an extra skip connection, i.e., simple identity mapping, to form a complete residual block (Figure 5). This residual block learns the necessary residual mapping for a given task and improves the capability of training considerably deeper networks. This method reduces training time and improves convergence speed, resulting in more accurate prediction.

The ResNet architecture has shown impressive performance not only on image classification benchmarks, such as CIFAR (Krizhevsky et al., 2009) and ImageNet, but also on object detection benchmarks such as MS COCO (Lin et al. 2014). Researchers have implemented numerous improvements in the structure of residual units to facilitate the learning of the network. For instance, Inception-ResNet (Längkvist et al., 2014) has been introduced by combining residual learning and inception blocks. It has been shown that training with residual connections significantly accelerates the training of inception networks. Merge-and-run mappings in residual units have been implemented to improve network performance by facilitating information flow in networks (Zhao et al., 2018). There are numerous variants of the ResNet architecture, which are based on the same concept but contain different numbers of layers, for instance, ResNet-18, ResNet-34, and ResNet-50 (He et al., 2016), where the number denotes the number of neural network layers (see Table 1). The CNN architecture used in this study is the original ResNet-34 architecture pretrained on the 1000-class ImageNet dataset to accelerate the convergence of training and maintain a suitable number of model parameters while ensuring a certain feature extraction effect.

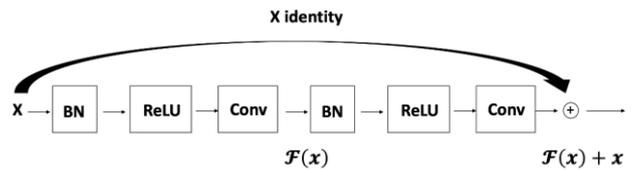

**Figure 5** Residual block in ResNet architecture.

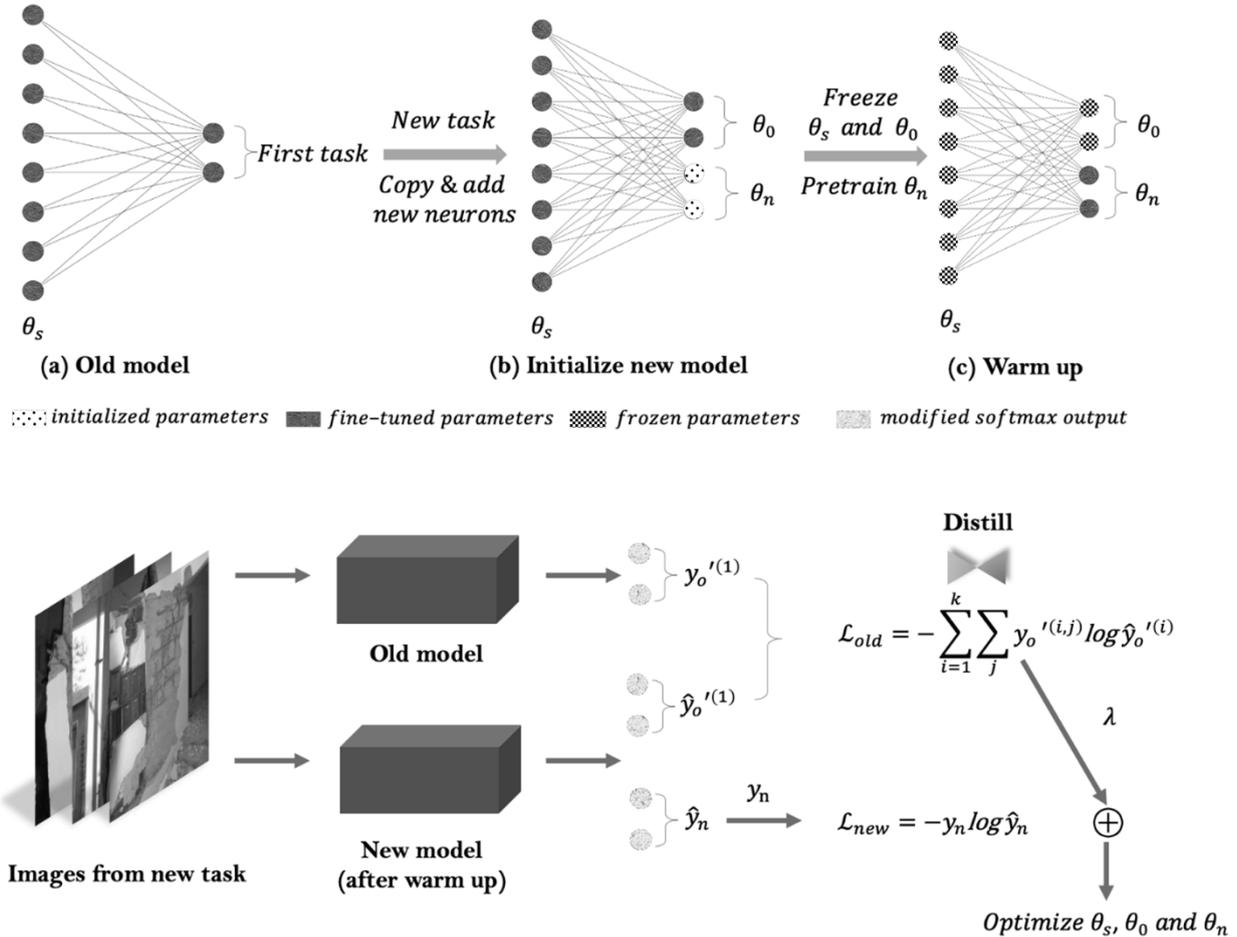

**Figure 6** CLDRM training framework.

### 4.1.2 Learning Without Forgetting

The L$w$F method was first developed by Li and Hoiem (2018), and has been adopted in the proposed CLDRM. There are three important notations in L$w$F, i.e., $\theta_s$, which denotes the parameters in the CNN shared across all tasks, $\theta_o$, which denotes the specific parameters learned from previous tasks, and $\theta_n$, which denotes the specific parameters that are learned in a new task.

In this study, ResNet-34 is utilized to implement multitask learning and $\theta_s$ (the parameters of layers before the tail of the fully connected layer) is initialized with the pretrained parameters of the 1000-class ImageNet dataset to accelerate the convergence of training. The features of images extracted from the first few layers of a CNN tend to be more generalized (Zeiler and Fergus, 2014), while the features extracted from the latter layers are more relevant to a specific dataset. Therefore, researchers tend to freeze the parameters of the first few layers after transferring the parameters trained on the dataset with richer features; only the parameters of the next few layers are fine-tuned (Yosinski et al., 2014). However, fine-tuning all parameters provides better performance in multitask learning. For this reason, $\theta_s$ is focused in the training process. In the tail part of the fully connected layer, fc-1000 in the original ResNet is replaced by a new linear layer with 512 input features and $m$ output features, where $m$ is the number of classes for the first task. Then, after the training process of the first task, $n$, which denotes the number of classes for the next new task, is appended to this fully connected layer. All newly appended parameters are initialized using Kaiming initialization (He et al., 2016).

Then, the size of the fully connected layer becomes $512 \times (m + n)$. $\theta_o$ and $\theta_n$ correspond to parameters $512 \times m$ and $512 \times n$, respectively.

The main strategy of this study is to add $\theta_n$ parameters for a new task within the same model and learn the parameters with $\theta_s$ and $\theta_o$ to improve the recognition accuracy for old and new tasks without using the data from old tasks (see Figure 6). Based on knowledge distillation (Hinton et al., 2015), Li and Hoiem (2018) referred to the old model that was trained for previous tasks as the teacher model and the new model with $\theta_n$ as the student model. The student model is allowed to learn from the teacher model to maintain the recognition accuracy for old tasks while learning a new task, with the same batch of images sent to both models.

The loss ($\mathcal{L}_{new}$) during the learning of a new task can be defined as the cross entropy between $y_n$ and $\hat{y}_n$, as expressed in Equation (1). $\hat{y}_n$ denotes the output corresponding to the new tasks of the new model, where $y_n$ is the ground truth.

$$\mathcal{L}_{new} = -y_n \log \hat{y}_n \qquad (1)$$

$y_o$ denotes the output (after softmax is applied separately for each old task) of the old model. $\hat{y}_o$ denotes the output corresponding to the old tasks of the new model. It is worth noting that the loss during the training for old tasks cannot be retrieved by directly adding the cross entropy calculated between $y_o$ and $\hat{y}_o$ for each old task; this results in poor prediction performance. For instance, if an image of a beam is used for prediction in the component type determination task, the old model may output a high probability for a beam and a relatively small probability for a column and a wall. However, the probability of a column may be considerably higher than a wall owing to its similarity to a beam. Both output probabilities are extremely small (i.e., $1 \times 10^{-4}$ and $1 \times 10^{-7}$ for a column and wall, respectively). Therefore, when information is passed to the student model for assisting with distinguishing the features of the two components, the information carried by the small probabilities tends to be lost. Thus, hyperparameter T (Equation (2)), which represents temperature (Hinton, Vinyals, and Dean, 2015), is introduced to amplify the information carried by the small probabilities. The modified probabilities of $y_o$ and $\hat{y}_o$ are given by Equation (2).

$$y_o'^{(i,j)} = \frac{(y_o^{(i,j)})^{\frac{1}{T}}}{\sum_j (y_o^{(i,j)})^{\frac{1}{T}}},$$

$$\hat{y}_o'^{(i,j)} = \frac{(\hat{y}_o^{(i,j)})^{1/T}}{\sum_j (\hat{y}_o^{(i,j)})^{1/T}} \qquad (2)$$

Superscript $i$ represents the $i^{th}$ old task in the previous training, and it varies between 1 to $k$, which represents the number of the old tasks that have been previously trained. Superscript $j$ represents the $j^{th}$ class in the $i^{th}$ old task. As T is set to be 2, the loss ($\mathcal{L}_{old}$) for an old task can be expressed Superscript $i$ represents the $i^{th}$ old task in the previous training, and it varies between 1 to $k$, which represents the number of the old tasks that have been previously trained. Superscript $j$ represents the $j^{th}$ class in the $i^{th}$ old task. As T is set to be 2, the loss ($\mathcal{L}_{old}$) for an old task can be expressed as Equation (3).

$$\mathcal{L}_{old} = -\sum_{i=1}^{k} H\left(y_o'^{(i)}, \hat{y}_o'^{(i)}\right)$$
$$= -\sum_{i=1}^{k} \sum_j y_o'^{(i,j)} \log \hat{y}_o'^{(i)} \qquad (3)$$

Overall, the training process can be concluded as follows:

- Step 1: The first task is trained using the typical training process in TL.
- Step 2: When a new task is added, a new model will be created and it inherits the $\theta_s$ and $\theta_o$ from the old model. The newly appended parameters $\theta_n$ will be initialized.
- Step 3: $\theta_s$ and $\theta_o$ are frozen, and $\theta_n$ is pretrained in the dataset of new task to convergence (warm up).
- Step 4: $\theta_s$ and $\theta_o$ are unfrozen, $\hat{y}_n$ and $\hat{y}_o'^{(i,j)}$ are output from the new model, and $y_o'^{(i,j)}$ is input from the old model using the same batch images of new task.
- Step 5: The total loss $\mathcal{L}_{total} = \mathcal{L}_{new} + \lambda \mathcal{L}_{old}$ is calculated with the ground truth $y_n$ and the outcome of the Step 3, namely, $\hat{y}_n, \hat{y}_o'^{(i,j)}, y_o'^{(i,j)}$. $\lambda$ is a hyperparameter that balances the learning tendency between the new task and the old tasks. In this study, $\lambda$ is set to 1.
- Step 6: $\theta_s, \theta_o$ and $\theta_n$ are simultaneously updated to minimize the total loss.

## 4.2 Comparative Evaluation

The performance of the proposed CLDRM framework is evaluated through experiments and compared to a conventional CNN model with four existing training methods, which are described below.

### 4.2.1 Feature Extraction

Feature extraction (Shin et al., 2016) is used to filter out the most critical and distinguishable information from redundant data. Extracted features are typically represented by feature vectors. In deep learning, the output of a certain layer (the last convolutional layer) can be extracted as the feature representation of input images and then used to classify images in the testing process (Donahue et al., 2014). Thus, in the multitask learning investigated in this study, $\theta_s$ and $\theta_o$ are defined as constants. Additionally, the features extracted from the last layer are used to determine $\theta_n$ through training so that training can be accelerated by neglecting backpropagation for $\theta_s$. However, the features extracted from the network are shared by all tasks. Such features are independent of any specific task; this slightly reduces classification accuracy.

### 4.2.2 Fine-Tuning

To overcome the abovementioned problem of feature extraction, $\theta_o$ is fixed and a low learning rate is applied for training $\theta_s$ and $\theta_n$ (Yosinski et al., 2014). In this manner, the network learns the features related to a new task. However, it may forget the features learned in the first task. The training conditions of the CLDRM and fine-tuning are quite similar but considerably different from the training conditions of the two methods mentioned below. Therefore, the CLDRM is mainly compared with fine-tuning in the following experiments.

### 4.2.3 Duplicate and Fine-Tuning

During the training for a new task, the relatively important parameters for the identification of the first task are partially modified. This causes the network to forget the features learned in the previous task. For this reason, the duplicate and fine-tuning method is applied to the network for each new task. Moreover, each network must be individually fine-tuned. It is worth noting that even though each network learns specific features from the images corresponding to a task, the number of parameters still increases exponentially.

### 4.2.4 Joint Training

Joint training simultaneously trains all parameters ($\theta_s$, $\theta_o$, and $\theta_n$) using all data from old and new tasks (Caruana, 1997).The network can extract and integrate the characteristics of each task. This may improve the accuracy for a few recognition tasks compared to the individual training of each task. However, storing all data is an issue and extremely expensive in terms of time and computation resources. In order words, whenever a new task is added, it must be trained again with the old tasks that have been previously trained on the network.

### 4.3 Model Training

This section describes the CLDRM training process, including the training and test sets, optimization methods, and hardware configuration. All tasks are performed on a workstation with the Intel(R) Xeon(R) E5-2678 v3 2.50 GHz CPU, 64.0 GB RAM, and the NVIDIA RTX2080TI-11G GPU.

### 4.3.1 Train and test set

The number of training and test sets used in the experiments for different recognition tasks are listed in Table 2. In the data preprocessing part, data augmentation (horizontal flip, vertical flip, rotation, color jitter, etc.) is implemented with a batch size of 64. However, a small proportion of the datasets in the structure component part contain useful environmental background information. Therefore, in most cases, at a certain angle of rotation of images, inclined beams may look like vertical columns and vice versa. Therefore, to prevent this, images are not rotated in the component type determination task.

**Table 2**

Number of images for the training and test sets for each task.

| Task | Damage Level | Spalling | Component | Damage Type |
|---|---|---|---|---|
| Training | 3776 | 4864 | 3968 | 1728 |
| Test | 663 | 820 | 693 | 328 |

### 4.3.2 Optimization

The results of the initial experiments for the training of continual learning tasks show that the Adam optimizer (Kingma and Ba, 2015) is relatively unstable and ineffective, which results in poor model performance. The recognition accuracy for an old task decreases significantly during the training of a new task. Thus, the Adam optimizer is not selected as the training method for the CLDRM. SGD (Nesterov, 1983) with momentum provides considerably better performance, and it is employed as the optimizer for the CLDRM. The learning rate of the SGD optimizer is $1 \times 10^{-3}$, the momentum is 0.9, and the weight decay rate is $4 \times 10^{-5}$. These parameter settings are adopted to pretrain $\theta_n$, and the model is trained for 40 epochs. When $\theta_s$, $\theta_o$, and $\theta_n$ are simultaneously trained, the learning rate is reduced to $1 \times 10^{-4}$ and the number of training epochs is increased to 60.

## 5 EXPERIMENTS AND RESULTS

The feasibility of the CLDRM is experimentally verified for four different recognition tasks, i.e., damage level evaluation, spalling condition check, component type determination, and damage type determination. The objective is to achieve accurate multitask damage recognition through a single neural network. However, typically, the recognition accuracy for a previous task decreases when new tasks are input to the neural network. The target is to minimize the decrease in the recognition accuracy for previous tasks so that the total accuracy is high. The influence of the following parameters on the model is examined: a) effects of different distillation temperature settings on the four proposed continuous-learning-based

**Table 3**

Comparison of accuracy for different recognition tasks trained by different methods.

| Task | Damage level | | Spalling Condition | | Component type | | Damage type Final (%) | Average of Final (%) |
|---|---|---|---|---|---|---|---|---|
| | Initial (%) | Final (%) | Initial (%) | Final (%) | Initial (%) | Initial (%) | | |
| *Feature Extraction* | 81.45 | 81.45 | 84.27 | 84.27 | 90.04 | 90.04 | 77.74 | 83.37 |
| *Fine-Tuning* | 90.05 | 83.41 | 88.41 | 84.15 | 95.67 | 92.21 | 92.07 | 87.96 |
| *Duplicate and Fine-Tuning* | 89.44 | 89.44 | 88.41 | 88.41 | 96.10 | 96.10 | 89.94 | 90.97 |
| *Joint Training* | 90.35 | 90.35 | 86.71 | 86.71 | 95.82 | 95.82 | 89.63 | 90.62 |
| *CLDRM* | 90.20 | 84.92 | 88.90 | 86.71 | 96.10 | 92.06 | 93.60 | 89.32 |

recognition tasks; b) influence of characteristic similarity on continual learning; c) influence of learning order on mixed similar and dissimilar tasks.

**5.1 Initial Study**

The initial experiment compares the accuracy of different recognition tasks using different training methods for model. Table 3 compares the accuracy of different recognition tasks and training methods. "Initial" refers to the best test accuracy for a task that is trained for the first time. "Final" refers to the test accuracy for this specific task when the training of all tasks is finished. The difference between "Initial" and "Final" for a task shows the decline in the recognition accuracy for that task during the entire training process. It is worth noting that the values of "Initial" and "Final" are the same for damage type determination because it is the last task to be trained. $\theta_s$ and $\theta_o$ are constant in feature extraction. Therefore, the values of "Initial" and "Final" are the same for all tasks in the case of this training method. In addition, there is no concept of "Initial" and "Final" in the duplicate and fine-tuning method and joint training method. However, for the convenience of comparison, "Final" is considered as equal to "Initial" for these methods.

As shown in Table 3, when the number of trained tasks increases, the decrease in the recognition accuracy for previous tasks is less in the case of the CLDRM compared to the fine-tuning and feature extraction methods. Moreover, compared to the other three methods, the CLDRM achieves the highest recognition accuracy (93.60%) for the final trained task (damage type determination). The Average of "Final" for the CLDRM is also relatively high.

The CLDRM achieves high accuracy primarily because it is capable of retaining the useful and similar characteristic features of the previous three tasks, which are beneficial for the damage type determination task. Hence, damage types are classified more accurately with less direct training data.

Similarly, the characteristics of the cracks caused by the ASR are quite similar to a certain type of spalling. Thus, the training of the spalling condition check task is beneficial for the ASR damage detection task. In the future, this notion can be implemented to improve the accuracy for recognition tasks where direct training alone cannot improve the accuracy.

In Table 3, the "Average of Final" indicator is used to measure the overall accuracy of different training methods. It is the average of the value of "Final" for each recognition task. The Average of "Final" for the CLDRM is 1.36% higher than the fine-tuning method. This difference increases with the number of tasks. The duplicate and fine-tuning method and joint training method have the highest Average of Final; this indicates that these methods perform the best. However, as mentioned in Section 4.2, the high accuracy of these methods comes at the high cost of a large number of parameters and data storage required during training.

The confusion matrices of the test prediction by the CLDRM in the four recognition tasks are shown in Figure 11. As shown in Figure 11 a), when the damage level is categorized into three classes ("undamaged," "minor damage," and "heavy damage"), the model successfully distinguishes between "heavy damage" and "minor damage" with high accuracy. However, in the same recognition task, "minor damage" is misclassified by the model as the "undamaged" in most cases. The main reason behind this might be that the characteristics of the damage in the "minor damage" dataset is not obvious and similar to that in the "undamaged" dataset. Furthermore, the damage in the dataset used for training the other recognition tasks is quite similar. For this reason, the characteristics used to distinguish between "minor damage" and "undamaged" might not be important for training the next recognition task and might not be considered in continual learning. This also explains the reason for the low accuracy of the CLDRM for the damage level evaluation task. A similar problem occurs in the case of the spalling condition check task; however, the decrease in recognition accuracy is small (Figure 11 b)). In the component type determination task, as there is a significant similarity between wall and beams, 14% wall images are misclassified as beams (Figure 11 c)). In addition, the CLDRM shows high accuracy for the damage type determination task because it learns the features of damage in the previous three tasks (Figure 11 d)).

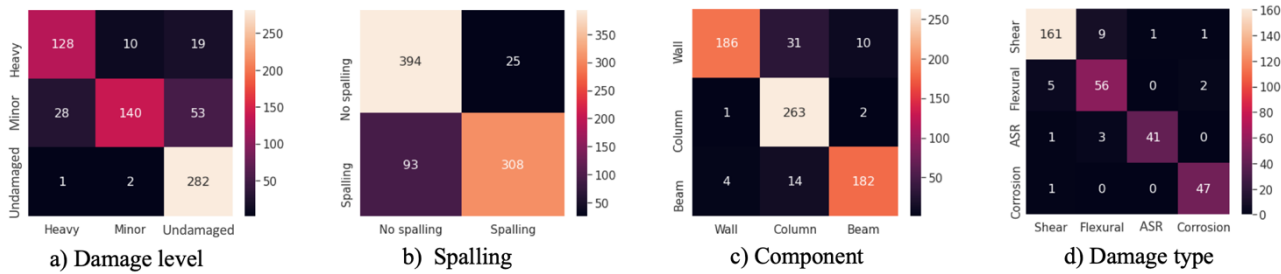

**Figure 7** Confusion matrices of test prediction by CLDRM for four recognition tasks.

## 5.2 Further Experiments

### 5.2.1 Distillation temperature

As mentioned in Section 4.1.2, when the new model distills the knowledge of old tasks from the old model, most detailed information is lost if the softmax output of the old model is relatively similar to one-hot encoding. Thus, Hinton et al. (2015) introduced the distillation temperature, $T$, to distill the characteristic information of low probability targets. The output of the old model is divided by $T$ before reaching the softmax layer. Therefore, a larger $T$ leads to a more uniform probability distribution, which implies that the information carried by a small probability target is amplified. In this study, $T$ is used to mitigate catastrophic forgetting in the CLDRM, i.e., to reduce the decrease in the recognition accuracy for old tasks while learning new tasks.

This section describes the effects of different temperature settings on the recognition accuracy for the first task. Component type determination is set as the first task, and spalling condition check is set as the second task with $T$ set to be $1, 2, 5, and$ 10. These tasks are selected owing to their distinct characteristics, which tend to provide better generalization compared to other tasks. The purpose of this experiment is to find the optimal value of the distillation temperature by observing the variations in the recognition accuracy for the first task during the training of the second task. A smaller decrease in the recognition accuracy for the first task implies a more suitable value of the distillation temperature. The first task is trained using is a typical process with no distillation operation. Therefore, the accuracies of the models trained for the first task at different temperatures are almost the same. The main variation in accuracy occurs during the training of the second task. As illustrated in Figure 12, the accuracy for the first task is obtained over 120 epochs of training, where the first 60 (0–59) and last 60 epochs represent the training of the first and second tasks, respectively. It is worth mentioning that there are only minor changes in accuracy during the training of the second task at different temperatures. This implies that the variations in the temperature have a negligible effect on the learning of the new task. Nonetheless, the temperature significantly impacts the knowledge distillation from the first task. As clearly shown in Figure 12, when $T = 1$, it is ineffective to use the normal SoftMax output from the old and new models to calculate the loss for retaining the recognition accuracy for the first task. However, accuracy increases at $T = 2$. The maximum accuracy is obtained at $T = 5$, and it is approximately 3–4% higher than the accuracy at $T = 1$. However, as T continues to increase more, accuracy tends to gradually decrease. For instance, the accuracy at $T = 10$ is less than that at $T = 1$. These results show that the appropriate distillation temperature improves the effect of knowledge distillation from old tasks. $T$ is set as 5 in the subsequent experiments to obtain the best accuracy.

### 5.2.2 Tasks with correlated Characteristics

This section describes the impact of the feature correlation between learning tasks on the performance of continual learning. When different tasks share numerous similar or same feature extractors, fine-tuning tends to perform better than joint training. The CLDRM effectively retains the knowledge of a previous task. Moreover, owing to the recurrence of related features, the training of the second task may even improve the recognition accuracy for old tasks in a few cases.

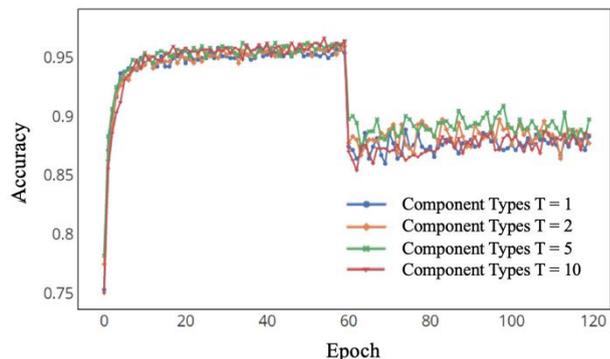

**Figure 8** Effect of different distillation temperatures on the recognition accuracy for the first task.

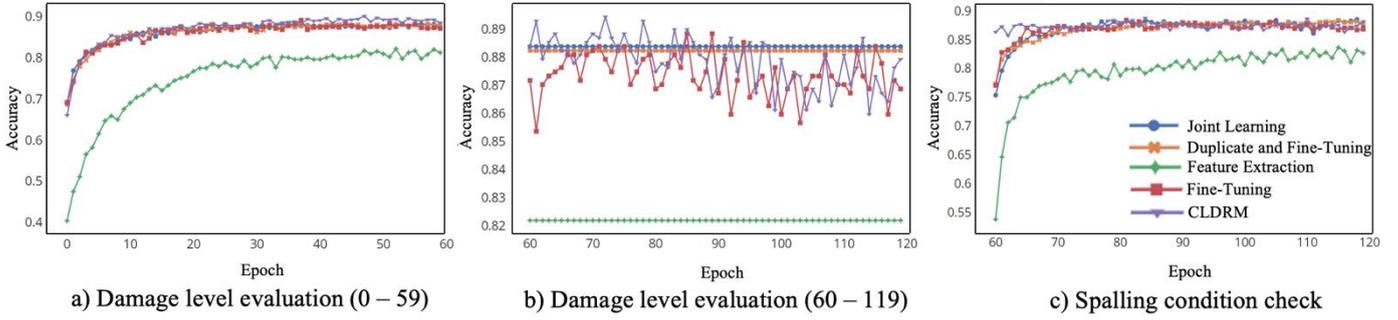

**Figure 11** Variation in recognition accuracy of the model trained with different methods for similar tasks.

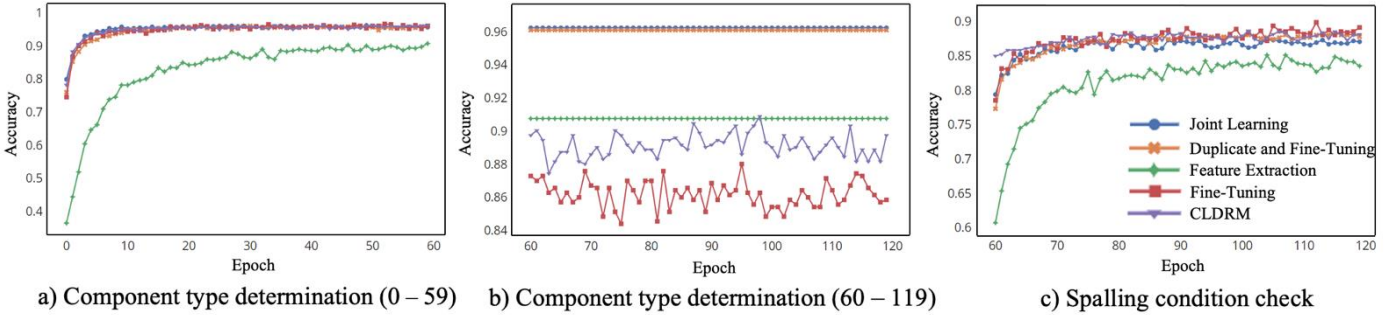

**Figure 9** Variations in recognition accuracy of the model trained with different methods on dissimilar tasks

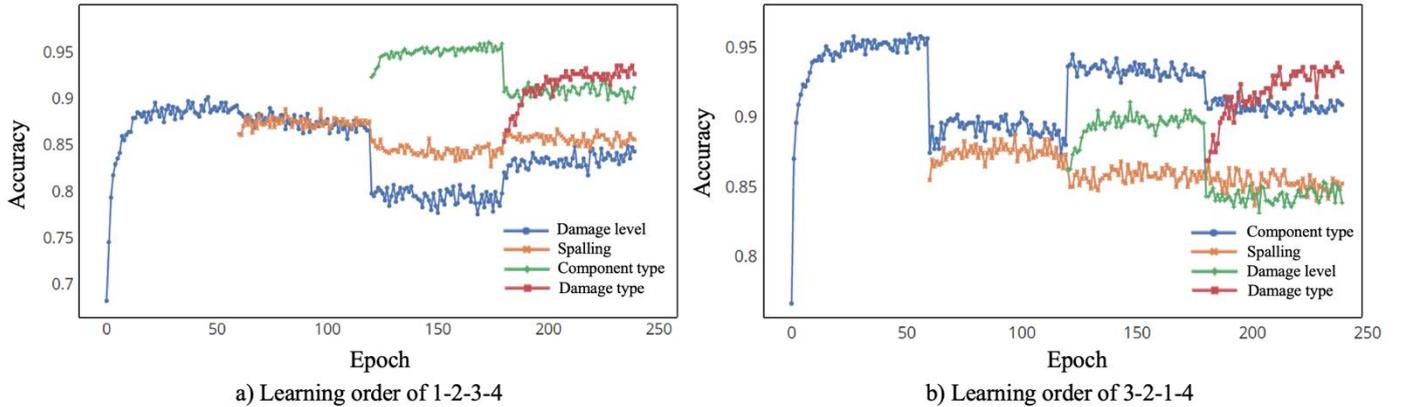

**Figure 10** Variation in recognition accuracy for different tasks and different learning orders.

First, tasks with similar characteristics are investigated. The damage level evaluation and spalling condition check tasks are used as the first and second learning tasks, respectively, owing to their characteristic similarities. Figure 13 a) and c) illustrates the variation in the recognition accuracy for the first (0–59 epochs) and second (60–120 epochs) tasks during training, respectively. Figure 13 b) shows the variation in the recognition accuracy for the old task during the training of the new task (60–120 epochs). Besides, the training performed using the joint training method and duplicate and fine-tuning method is only for 60 epochs each. During the training of the new task, $\theta_s$ and $\theta_o$ are constant for the feature extraction method (Figure 13 a) and b)). The accuracies of the three methods are shown in Figure 13 b)). The results show that the overall performance of the CLDRM is better than the fine-tuning method. In addition, the CLDRM significantly outperforms the feature extraction method. The CLDRM utilizes the correlation between the

**Table 4**

Recognition accuracy of CLDRM for different learning orders.

| Task | Damage level | | Spalling Condition | | Component type | | Damage type | Average of |
| | Initial (%) | Final (%) | Initial (%) | Final (%) | Initial (%) | Final (%) | Final (%) | Final (%) |
|---|---|---|---|---|---|---|---|---|
| *1-2-3-4* | 90.20 | 84.92 | 88.90 | 86.71 | 96.10 | 92.06 | 93.60 | 89.32 |
| *1-3-2-4* | 89.74 | 84.01 | 88.78 | 87.32 | 96.54 | 91.34 | 93.29 | 88.99 |
| *2-1-3-4* | 90.65 | 85.67 | 88.05 | 85.00 | 96.25 | 91.63 | 93.90 | 89.05 |
| *2-3-1-4* | 91.10 | 84.92 | 88.41 | 85.37 | 96.83 | 91.77 | 93.29 | 88.83 |
| *3-1-2-4* | 89.89 | 84.62 | 88.17 | 88.29 | 96.10 | 91.20 | 93.90 | 89.50 |
| *3-2-1-4* | 91.10 | 85.37 | 88.78 | 86.83 | 95.96 | 91.77 | 93.90 | 89.46 |

features of tasks; thus, there is a decrease of only 1–2% in the recognition accuracy for the old tasks during the training of the new tasks. As Li and Hoiem (2018) reported, the joint training method is an upper bound on the accuracy for other state-of-the-art methods.

However, the performance of joint training is worse than the CLDRM in a few cases. For instance, if spalling images are used for training the spalling condition check task before the heavy damage determination task, the output neurons for the damage level evaluation task can utilize a large number of features from the neurons for detecting spalling. This is owing to the similarity of the feature extractor as features two tasks shared by the two tasks. Therefore, the probability distribution of the output is relatively similar to the probability distribution of the spalling condition check task. However, in the backpropagation step, an increase in the probability of spalling condition check decreases the output probability of damage level evaluation neurons, and the extent of the decrease is unpredictable. Conversely, if the similarity between the features of multiple recognition tasks is low, then the neurons activated in feature extraction for the output neurons of one task are not too much overlap with those used by other tasks. Thus, the backpropagation during the training of other tasks does not strongly affect the neurons. In this case, the performance of joint training is considerably better than the CLDRM or fine-tuning method.

To investigate tasks with dissimilar characteristics, the component type determination and spalling condition check tasks are used as the first and second tasks, respectively, owing to the difference between their characteristics. Figure 14 a) shows the variation in the accuracy for the first task in the first 60 epochs, and Figure 14 c) illustrates the variation in the accuracy for the second task in the last 60 epochs of training. Figure 14 b) demonstrates the effects of the training of the second task on the accuracy for the first task. When the characteristics of different tasks are not similar, the performance of the joint training method is better than the CLDRM, as predicted in Section 5.2.2. This observation provides useful guidance for the subsequent adjustments of the learning order of the CLDRM; this is discussed in detail in the next section. Moreover, as seen from Figure 14 b), the CLDRM can retain more knowledge and information of old tasks compared to the fine-tuning method. In addition, for the fine-tuning method, the changes in the weights of the important neurons for the old tasks cannot be controlled during the continual learning of new tasks. Therefore, catastrophic forgetting occurs as the number of tasks increases. On the contrary, the model with the CLDRM not only retains the important features of old tasks but also increases the recognition accuracy for these tasks during the training of new tasks, which is related to the learning order. In general, compared to the continual learning of similar tasks, the recognition accuracy of the model trained with the CLDRM on dissimilar tasks tends to experienced larger decrease.

### 5.2.3 Learning Order

This section describes the effect of the learning order of different tasks on the recognition accuracy of the CLDRM by mixing the feature-related and feature-unrelated tasks. The purpose of this experiment is to find whether the learning order of these four tasks has a significant impact on the final accuracy for a particular task. Owing to the advantage of feature fusion from different tasks in continuous multitask learning, this experiment will examine whether it is possible for its recognition accuracy to be higher than the one that was individually trained using fine-tuning method. In other words, we examine whether the knowledge of the spalling condition check and component type determination tasks helps improve the accuracy of the damage type determination task. The damage type determination task is selected as the last task to be trained, and the learning order of the other three tasks is arbitrarily changed. The damage level evaluation, spalling condition check, component type determination, and damage type determination tasks are denoted by numbers 1, 2, 3, and 4, respectively. For instance, the learning order of the damage level evaluation, spalling condition check, component type determination, and damage type determination tasks is denoted as *1-2-3-4*.

Table 4 shows the recognition accuracy of the CLDRM for all tasks for different learning orders. The CLDRM shows excellent performance in maintaining the knowledge of old tasks. The decrease in the recognition accuracy for old tasks during the training of new tasks is less than 6%. Additionally, the learning order of *3-1-2-4* shows the highest performance

**Table 5**

Recognition accuracy of fine-tuning method for different learning orders.

| Task | Damage level Initial (%) | Damage level Final (%) | Spalling Condition Initial (%) | Spalling Condition Final (%) | Component type Initial (%) | Component type Final (%) | Damage type Final (%) | Average of Final (%) |
|------|------|------|------|------|------|------|------|------|
| *1-2-3-4* | 90.05 | 83.41 | 88.41 | 84.15 | 95.67 | 92.21 | 92.07 | 87.96 |
| *1-3-2-4* | 89.59 | 81.45 | 88.90 | 83.22 | 95.82 | 91.20 | 90.55 | 86.61 |
| *2-1-3-4* | 90.65 | 85.67 | 88.05 | 84.00 | 96.25 | 87.63 | 91.90 | 87.30 |
| *2-3-1-4* | 91.10 | 84.92 | 88.41 | 83.37 | 96.83 | 88.77 | 91.29 | 87.08 |
| *3-1-2-4* | 89.89 | 84.62 | 88.17 | 83.29 | 96.10 | 86.20 | 91.90 | 86.50 |
| *3-2-1-4* | 91.10 | 84.92 | 88.17 | 84.02 | 95.96 | 85.28 | 92.99 | 86.80 |

**Table 6**

Evaluation of different training methods used for the recognition tasks.

| Methods | Feature Extraction | Fine-Tuning | Duplicate and Fine-Tuning | Joint Training | CLDRM |
|---------|------|------|------|------|------|
| Old tasks performance | Medium | Medium | Good | Good | Good |
| New task performance | Medium | Good | Good | Good | Good |
| Training time | Fast | Medium | Medium | Medium | Medium |
| Robust to the learning order | Good | Medium | Good | Good | Medium |
| Require previous datasets | No | No | No | Yes | No |

among these six learning orders, with an average accuracy of 89.50% for the four tasks. However, this does not imply that the recognition accuracy for every task is the highest in this order. For example, the value of "Final" for the damage type determination task is the highest for *2-1-3-4* instead of *3-1-2-4*, even though it is the second task in both cases.

Figure 15 compares the recognition accuracy for the four tasks between two different learning orders, i.e., *1-2-3-4* and *3-2-1-4*. During the training for the final task (damage type determination), the accuracies for the first task in *1-2-3-4* (damage level evaluation) and *3-2-1-4* (component type determination) increase by almost 4% and 6%, respectively. This occurs under the condition that the same characteristics are adopted for training previous and new tasks. However, in the training for a new task, the accuracy for an old task first decreases slightly and then increases. For instance, in *3-2-1-4*, the accuracy for the first task significantly decreases by 8% when new tasks (spalling condition check) are added. If the training is terminated after the third task instead of the fourth task, then the influence of the learning order on accuracy is relatively large.

The accuracies of the feature extraction, duplicate and fine-tuning, and joint training methods are not affected by the learning order. Table 5 shows the recognition accuracy of the fine-tuning method for different learning orders.

The overall recognition accuracy of the fine-tuning method is approximately 1–2% lower than the CLDRM. Regardless of the learning order, the accuracy for the first task learned by the fine-tuning method decreases significantly by an average of 8% after learning the fourth task, as compared to only 5% in the CLDRM.

**5.3 Model Evaluation**

The performance of each training method is evaluated on the test set using the indicators listed in Table 6. The feature extraction method requires the least training time but provides mediocre performance for new tasks. The duplicate and fine-tuning method provide good performance on old and new tasks; however, the prediction time of the corresponding model is quite high. The joint training method provides superior performance for old and new tasks but requires a large amount of data storage while training. On the contrary, the CLDRM not only provides high recognition accuracy and speed for old and new tasks but also requires less data storage and parameters during training. The results provided in the table are for a generalized evaluation purpose only. These results are not only affected by whether tasks are related but also by specific datasets and application scenarios. In addition, one of the main challenges is the amount of data storage or computing power required for amount of data storage or computing power required for model training. Currently, model training and its application are mainly performed on a powerful desktop station or laptop. Thus, storing a large number of model parameters and datasets might not be a major issue owing to the rapid development of hard drives, CPUs, and GPUs. Therefore, in this scenario, it is acceptable to apply the joint training method or duplicate and fine-tuning method for achieving rapid training and prediction. However, if a model must be deployed on mobile devices, such as a mobile phone or a UAV (Unmanned Aerial Vehicle), it will be considerably challenging to achieve effective real-time prediction because the storage capacity and computing resources of these devices are generally limited and they cannot store a large number of model parameters. Therefore, the CLDRM is a better method in this case.

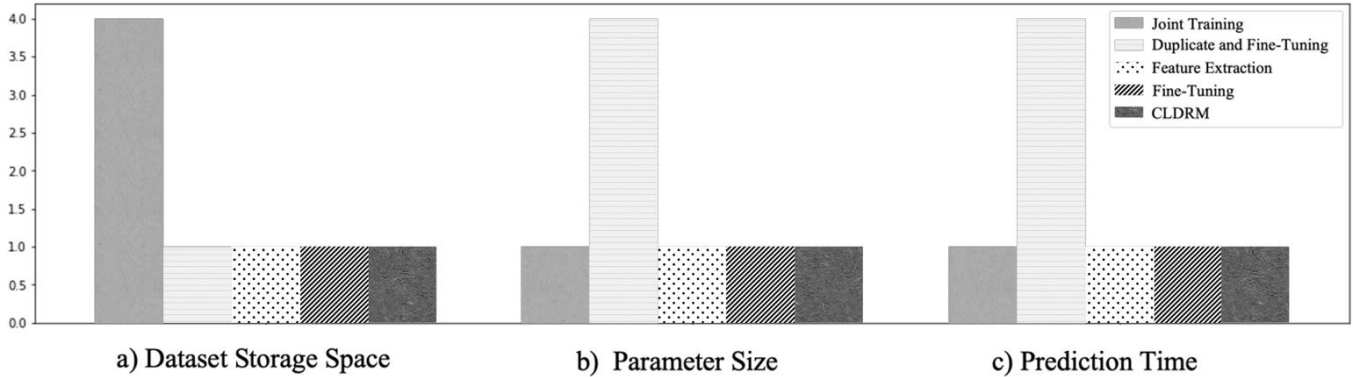

**Figure 12** The relative evaluation comparison on three indicators of computational cost
when taking the performance of CLDRM as the unit.

As it has shown the different performance for the five methods on the prediction accuracy on different datasets, and it demonstrates that the method "Joint training" and "Duplicate and Fine-Tuning" present a bit higher prediction accuracy in some cases. However, the trade-off is that they will require more dataset storage and parameters to maintain that performance. Besides, the prediction time for duplicate and fine-tuning will gradually become unacceptable as the number of tasks grows. For better demonstrating the cost of them, the relative evaluation on three indicators in four-task continual learning experiments are shown in Figure 12, which take the performance of CLDRM as the unit. Taking Table 4 into consideration, the CLDRM keeps the best balance between prediction accuracy and computational cost.

## 6. CONCLUSION AND FUTURE WORKS

In order to meet the requirement of multi-damage recognition in engineering practice, this study proposed a new deep CNN framework for the damage detection of RC structures, namely, the CLDRM, by combining the state-of-the-art L$w$F with the ResNet 34 architecture. The newly proposed approach not only possesses high prediction accuracy, but also have the advantage of computational economy. Deep CNN networks with traditional training methods, i.e., feature extraction, fine-tuning, duplicate and fine-tuning, as well as joint training were also investigated and compared with the new method. Three different experiments for four recognition tasks, including damage level, spalling check, component type and damage type determination were designed based on this training method to explore the optimal model parameters and applicable scenarios in damage recognition. The conclusions are as follows:

- Compared to conventional neural network models, the CLDRM can continuously train a model for multiple recognition tasks, without losing the prediction accuracy of old tasks. Additionally, the CLDRM provides robust performance with higher recognition accuracy and faster prediction for old and new tasks. It is achieved by optimized the model with less parameters and data storage. To be more specific, it will need just a quarter of the parameters needed for duplicate and fine-tuning method in most cases.

- In practical applications, the appropriate distillation temperature depends on the characteristics of a trained dataset. An appropriate distillation temperature can improve the recognition accuracy of the CLDRM. However, recognition accuracy decreases when the temperature is extremely high.

- The CLDRM is more suitable for feature-related multitask learning. The recognition accuracy for old tasks decreases negligibly when learning new tasks. Conversely, there might be a significant decrease in the accuracy for old tasks in feature-unrelated multitask learning. However, the CLRDM still maintains an accuracy of more than 80%.

- The learning order influences the CLDRM. It is important to determine the appropriate learning order of tasks and select when to end the learning process for achieving better performance. To improve the recognition accuracy of the CLDRM for a particular task after fine-tuning, this task must be trained as the last task. This process promotes feature fusion to help improve the recognition accuracy for the final task.

Owing to an increase in the demand for structural recognition tasks and the number of new datasets available for application in structural damage inspection, it is crucial to rapidly train and deploy new recognition tasks in a model with reliable performance without retraining all tasks. In this aspect, the performance of the CLDRM is better than the conventional training methods. Unlike TL, continual learning is a relatively new training technique in deep learning, and it has not yet been widely applied in civil engineering applications. Several extensions of continual learning can be explored in future, as follows:

- In terms of parameter updates during model training, certain restrictions can be placed on the weight updates of the neurons that are relatively beneficial for maintaining damage characteristics to prevent catastrophic forgetting(Kirkpatrick et al., 2017).
- A GAN (Generative Adversarial Network) model (Goodfellow et al., 2014) can be simultaneously trained with a particular task to maintain the memory of the dataset used in this training. Thus, when new tasks are added in the future, the past data collected by the GAN model can be directly used for training. This is equivalent to improving the training efficiency of joint training.
- Combining long-term memory and short-term memory, such as LSTM (Long Short Term Memory) (Hochreiter et al. 1997), for model training can be further explored to alleviate the adverse effects of the gradual forgetting problem of continual learning (Gepperth and Karaoguz, 2016). In this manner, the model can preserve the recognition accuracy for a previously learned task and simultaneously improve the accuracy for a new recognition task because it is extremely capable of retaining information about the characteristics of previous learning tasks.
- The CLDRM has the potential to be used for portable devices owing to its small number of parameters and ability to rapidly and accurately learn new tasks. Once the model achieves stable performance with acceptable recognition accuracy, its algorithms can be programmed into portable hardware or as mobile software applications (Li and Zhao, 2019) and combined with modern drone/robot-based detection (Montero et al. 2017) to help perform real-time structural damage recognition and evaluation in the post-disaster structural inspection.

**ACKNOWLEDGMENTS**

The authors would like to gratefully acknowledge the support from the National Key R&D Program of China (2018YFE0125400) and the National Natural Science Foundation of China (U1709216), which made the research possible. The project was also funded by Centre for Balance Architecture, Zhejiang University. In addition, the authors also need to acknowledge that the first author and the second author have equivalent contribution to the paper.